\documentclass[conference]{IEEEtran}
\IEEEoverridecommandlockouts
\usepackage{cite}
\usepackage{amsmath,amssymb,amsfonts}
\usepackage{algorithmic}
\usepackage{graphicx}
\usepackage{textcomp}
\usepackage{xcolor}
\usepackage{comment}
\usepackage{todonotes}
\usepackage[caption=false]{subfig}
\usepackage{url}

\IEEEpubid{\begin{minipage}{\textwidth}\ \\[12pt]
978-1-7281-1884-0/19/\$31.00 \copyright 2019 IEEE
\end{minipage}}

\def\BibTeX{{\rm B\kern-.05em{\sc i\kern-.025em b}\kern-.08em
    T\kern-.1667em\lower.7ex\hbox{E}\kern-.125emX}}
\begin{document}

\title{Your Gameplay Says It All:\\ Modelling Motivation in \emph{Tom Clancy's The Division}
}

\author{
\IEEEauthorblockN{David Melhart\IEEEauthorrefmark{1},
Ahmad Azadvar\IEEEauthorrefmark{2},
Alessandro Canossa\IEEEauthorrefmark{2},
Antonios Liapis\IEEEauthorrefmark{1} and
Georgios N. Yannakakis\IEEEauthorrefmark{1}}

\IEEEauthorrefmark{1}\textit{Institute of Digital Games, University of Malta}
    \\\{david.melhart, antonios.liapis, georgios.yannakakis\}@um.edu.mt\\
\IEEEauthorrefmark{2}\textit{Massive Entertainment a Ubisoft Studio}
    \\\{ahmad.azadvar, alessandro.canossa\}@massive.se}

\maketitle

\begin{abstract}
Is it possible to predict the motivation of players just by observing their gameplay data? Even if so, how should we measure motivation in the first place? To address the above questions, on the one end, we collect a large dataset of gameplay data from players of the popular game \emph{Tom Clancy's The Division}. On the other end, we ask them to report their levels of \emph{competence}, \emph{autonomy}, \emph{relatedness} and \emph{presence} using the \emph{Ubisoft Perceived Experience Questionnaire}. After processing the survey responses in an ordinal fashion we employ preference learning methods based on support vector machines to infer the mapping between gameplay and the reported four motivation factors. Our key findings suggest that gameplay features are strong predictors of player motivation as the best obtained models reach accuracies of near certainty, from $92\%$ up to $94\%$ on unseen players.
\end{abstract}


\section{Introduction}

Fostering long term player engagement has been a traditional core challenge of game design and development. Players who are self-motivated to return to the game and keep playing are critical for a game's success \cite{rigby2011glued,yannakakis2014emotion,yannakakis2018artificial}. The central role of \emph{motivation} for the design of games, and the experiences they elicit, has been highlighted by a growing number of studies which adopt psychological theories of motivation within games \cite{canossa2013give,canossa2015your,melhart2018model}.
Such studies, however, follow a top-down integration of phenomenological models of motivation, which aim to identify and explain stereotypical player behaviour. 
Over the last decade, games user research and industry-based game testing has shifted its focus towards quantitative approaches to understand player behaviour and experience, based on \emph{player analytics} \cite{drachen2009player,makarovych2018like}. These approaches focus mainly on either clustering players based on their behavioural patterns or predicting objectively-defined aspects of their gameplay behaviour (e.g. churn prediction) \cite{yannakakis2018artificial}. 
{Despite these efforts, the majority of approaches that aim to capture aspects of player experience, such as engagement or motivation, remain qualitative in their analysis---even when quantitative data is involved---due to the complexity of measuring subjective notions of user experience in games \cite{el2016game}.}

Motivated by the lack of quantitative studies on the relationship between motivation and play, in this paper we introduce a data-driven \emph{player modelling} approach \cite{yannakakis2013player} by assuming there is an unknown underlying function between what a player does in the game \emph{behaviourally}---as manifested through her gameplay data---and her motivation. In particular, we assume that solely behavioural data from a player's gameplay would yield accurate predictors of motivation in games. To define motivation we rely theoretically on \emph{Self Determination Theory} \cite{ryan2000self} and examine four core factors: \emph{competence}, \emph{autonomy}, \emph{relatedness} and \emph{presence}, the latter of which is often associated with the theory in the domain of videogames \cite{rigby2011glued}. 

To study motivation quantitatively we are grounded in recent developments in motivation measurement tools, namely the \emph{Ubisoft Perceived Experience Questionnaire} (UPEQ) \cite{azadvar2018upeq}, which was developed as a game-specific tool to observe player motivation. To infer the relationship between player motivation and gameplay we collect data from more than 400 players of \emph{Tom Clancy's The Division} (Ubisoft, 2016). We process and aggregate this data and collect surveys on the players' motivation in relation to the game independently. We use the UPEQ questionnaire to measure players' general levels of \emph{competence,} \emph{autonomy}, \emph{relatedness} and \emph{presence} in the game.
Given the subjective nature of the reported notions we adopt a \emph{second-order} data processing approach \cite{yannakakis2018ordinal} and we process the reported UPEQ Likert-scale values of the players as ordinal data, and not as scores. We then apply simple statistical rank-based models and \emph{preference learning} \cite{furnkranz2011preference} methods based on support vector machines to infer the function between gameplay and reported factors of motivation. Our results suggest that factors of reported motivation can be predicted with high accuracy just relying on a few high-level gameplay features. In particular, the nonlinear machine learned models manage to predict the four motivation factors of unseen players with around $80\%$ average accuracy; the best models are reaching an accuracy of $94\%$ for \emph{competence} and \emph{autonomy} and $92\%$ for \emph{relatedness} and \emph{presence}. The obtained results add to the existing evidence for the benefits of ordinal data processing on subjectively-defined notions \cite{yannakakis2018ordinal} and validate that motivation can be accurately captured in the examined game based only on behavioural high-level data of playing.

This paper is novel in a number of ways. First, this is the first time player motivation (as in \emph{self-determination}) is modelled computationally only through gameplay data in games. Second, we introduce a \emph{second-order} \cite{yannakakis2018ordinal} methodology for treating Likert-scale scores which are used frequently in game testing and games user research at large. The ordinal approach we adopt compares the subjective scores of all players with each other and hence generates combinatorially very large datasets based only on small sets of participants. The approach is also effective in eliminating reporting biases of respondents, thereby better approximating the ground truth of reported motivation. Third, we model aspects of player motivation using preference learning based solely on a small number of key gameplay features. Finally, for the first time we evaluate the method in \emph{Tom Clancy's The Division} (Ubisoft, 2016) on a large set of
players and the predictive capacity of the motivation models for this game reach near certainty (i.e., over $80\%$ of average accuracy).

\section{Background: Measuring and Modelling Motivation}

This section gives an overview of related research on player modelling and motivation studies in games (Section \ref{sec:pm}) and then it introduces the fundamental principles of \emph{Self Determination Theory} (SDT) and the notion of \emph{presence} within the framework of UPEQ (Section \ref{sec:upeq}). The section ends with a discussion on the strengths of preference learning for modelling subjectively-defined psychological constructs such as motivation (Section \ref{sec:ordinal}).

\subsection{Player Modelling and Player Motivation}
\label{sec:pm}

Player modelling is one of the primary foci of AI research in the field of videogames \cite{yannakakis2018artificial}. It is generally concerned with the prediction of player behaviour or other cognitive and affective processes. Among their many uses, player models can inform and shape a game's monetization strategy, they can be directly applied as drivers of personalised content generation \cite{yannakakis2011experience,shaker2010towards}, and they can equip agents for believable user testing \cite{holmgaard2014evolving}.

Player modelling often relies on \emph{clustering} or \emph{prediction} \cite{yannakakis2018artificial}. Clustering is based on unsupervised learning methods with the aim to cluster players within groups of common behavioural patterns; approaches include $k$-means, self-organising maps \cite{drachen2009player}, matrix factorisation and archetypal analysis \cite{bauckhage2015clustering,drachen2012guns}, and sequence mining \cite{martinez2011mining,wallner2015sequential,makarovych2018like}. Prediction uses supervised learning to predict patterns of playing such as completion time \cite{mahlmann2010predicting} and churn \cite{runge2014churn,perianez2016churn,viljanen2018playtime}. 
Predictive models can predict a player's \emph{behaviour} (i.e., \emph{what would a player do?}) \cite{bakkes2012player} or the game \emph{experience} (i.e., \emph{what would a player feel?}) \cite{yannakakis2018artificial}. This type of modelling often uses behavioural data but also other modalities of player input such as physiological data \cite{martinez2011generic,georges2018physiology}, independently or in a multimodal fashion \cite{shaker2013fusing,camilleri2017towards}.

In the literature, motivation applied to games is built primarily on theoretical models that inform the design of experimental protocols instead of defining target outputs for predictive modelling. Indicatively, Borbora et al. \cite{borbora2011churn} use Yee's player motivation \cite{yee2006motivations} typology to enhance churn prediction, while Shim et al. \cite{shim2011exploratory} use motivational survey data to model player enjoyment. Birk et al. \cite{birk2015modeling} also incorporated motivational survey data along with top-down personality and player profiles to model enjoyment and effort. Although Birk et al. include high-level telemetry of progression (game stage) as a control variable in their experiments, they rely mainly on survey data. While the above studies use motivational survey data as top-down domain knowledge that forms the input of enjoyment models, this paper instead focuses on predicting motivation based on game metrics alone. Similarly to this study, Canossa et al. \cite{canossa2013give} applied regression models of telemetry to predict psychological survey data. However, that study was not aimed at player motivation modelling but instead used the Reiss Motivational Profile \cite{reiss1998toward} as a tool for personality profiling based on personal motivational drives. This paper focuses on the structure of motivation by considering four of its core manifestations, but our models do not rely on personality profiles of the players; instead, we try to infer the unknown mapping between gameplay features and motivation.

\subsection{From Theory to Measures of Motivation}
\label{sec:upeq}

Self-determination theory (SDT) is a well-established positive psychology theory of the facilitation of motivation \cite{deci1985intrinsic} which has been adopted in a wide variety of domains, including videogames \cite{ryan2006motivational,przybylski2010motivational,rigby2011glued,melhart2018model}. SDT distinguishes between an \emph{intrinsic} and \emph{extrinsic locus of causality} behind motivation \cite{ryan1989perceived}. The latter is facilitated by external or internal rewards, pressures, and expectations, while the former is based on the intrinsic properties of the activity itself, namely how well it can support \emph{competence}, \emph{autonomy}, and \emph{relatedness}. While  videogames include pressures and rewards which can promote extrinsic motivation \cite{deci1999meta}, they are generally regarded as good facilitators of intrinsic motivation \cite{przybylski2010motivational,melhart2018model}. Ryan et al. \cite{ryan2006motivational} describe the basic psychological needs (and \emph{presence}) underlying intrinsic motivation in videogames as:

\subsubsection{Competence} a sense of accomplishment and a desire for the mastery of an action. It is tied to self-efficacy and a sense of meaningful progression. It is supported through the interactions that players must master to complete the game, but not completion in itself.

\subsubsection{Autonomy} a sense of control and a desire for self-determined action. It manifests through meaningful choices, tactics, and strategic decisions that players can take. It is supported through rules and mechanics that structure the play experience but allow for a high degree of freedom and meaningfully different outcomes.

\subsubsection{Relatedness} a sense of belonging and a desire to connect and interact with others. It manifests through interactions with other players and believable computer agents. It is supported by multilayer interactions, believable and rich non-player characters, narrative design, and even interactions with other players outside the game.

\subsubsection{Presence} the feeling of a mediated experience is a main facilitator of both \emph{competence} and \emph{autonomy}. It can be viewed as having physical, emotional, and narrative components \cite{rigby2011glued,ryan2006motivational}. While it is not considered as one of the basic  psychological needs, presence or \emph{immersion}  can be a driving force behind gameplay motivation \cite{lombard1997heart,calleja2011game,melhart2018model}, and it is measured by both the \emph{Player Experience of Need Satisfaction} Questionnaire \cite{ryan2006motivational} and UPEQ \cite{azadvar2018upeq}.

It is important to note that the above factors are not contributing equally to the formulation of intrinsic motivation; while \emph{competence} or \emph{relatedness} are regarded as the core catalysts, \emph{autonomy} generally plays a supporting role in the facilitation of motivation. Nevertheless, in absence of \emph{autonomy}, motivation can only be considered \emph{introjected} or compulsive \cite{ryan2000intrinsic}. Within games the main drive of intrinsic motivation is generally \emph{competence} because of how the activity is structured, while \emph{relatedness} contributes to enhancing the experience \cite{rigby2011glued}.

This paper measures the four factors of SDT as affected by the gameplay experience via UPEQ, a game-tailored questionnaire developed by researchers at \emph{Massive Entertainment} \cite{azadvar2018upeq} to predict gameplay outcomes relevant for industry designers and stakeholders. 
{UPEQ measures the aforementioned factors of SDT through a 24 item survey using 5-point Likert-scales. From the 24 items, 21 measure the basic psychological needs of \emph{competence, autonomy}, and \emph{relatedness}, while 3 additional measure \emph{presence}.}
Earlier work \cite{azadvar2018upeq} has demonstrated that UPEQ is able to predict playtime, money spent on the game, and group playtime based on measured factors of SDT. UPEQ also addresses the limitations of prior domain-specific SDT questionnaires, such as the \emph{Game Engagement Questionnaire} \cite{brockmyer2009development}, \emph{BrainHex} \cite{nacke2014brainhex}, and the \emph{Player Experience of Need Satisfaction} \cite{ryan2006motivational}, by focusing on the adaptation of the \emph{Basic Need Satisfaction Scale(s)} \cite{chen2015basic} into a survey specific to videogame play. The result is a reliable and consistent assessment tool with a strong theoretical foundation in SDT.

\subsection{The Ordinal Nature of Motivation \label{sec:ordinal}}

This study uses preference learning (PL) methods as a predictive player model, due to the assumption of an ordinal nature of player experience. PL focuses on the \emph{differences} between occurrences instead of their \emph{absolute values} \cite{yannakakis2017ordinal,yannakakis2018ordinal}. This approach falls much closer to the players' cognitive processes---e.g. \emph{anchoring-bias} \cite{damasio1994descartes,seymour2008anchors}, \emph{adaptation} \cite{helson1964adaptation}, \emph{habituation} \cite{solomon1974opponent}, and other \emph{recency-effects} \cite{erk2003emotional}---that help them evaluate their own experience internally. 

There is growing evidence supporting the strength of preference learning for modelling emotions and user experiences both on a conceptual \cite{yannakakis2017ordinal,yannakakis2018ordinal} and a technical basis \cite{martinez2014don,yannakakis2015grounding,yannakakis2015ratings,melhart2018study}. Conceptually, treating subject-defined ground truth data as ordinal variables brings the representation of data closer to the players' underlying true attitudes \cite{yannakakis2017ordinal}. On the technical side, studies have compared processing of affective annotations as both ratings (e.g., Likert items) and rankings and found that 1) \emph{first-order} data processing (i.e. ranks) yields higher reliability and inter-rater agreement and 2) \emph{second-order} processing of the absolute rating values was also beneficial with regards to both reliability and validity \cite{yannakakis2015grounding,yannakakis2015ratings,yannakakis2018ordinal}, even when applied across different games \cite{camilleri2017towards} or dissimilar affective corpora \cite{melhart2018study}.

Given the above theoretical framework on the ordinal nature of experience and the large body of recent empirical evidence on the benefits of the ordinal modelling approach \cite{yannakakis2018ordinal}, in this paper we view player motivation as an emotional construct \cite{rigby2011glued} with ordinal properties. We compare player feedback on relative grounds and use PL to model the ranking between reported motivation in players as measured by the factors of UPEQ. While earlier work focused on the internal validity of the survey data \cite{azadvar2018upeq}, this paper considers the UPEQ scores as the underlying \emph{ground truth} to be predicted. After acquiring a general score for all the measured factors for each participant, we analyse and model the data as ordinal values, thereby following a \emph{second-order} modelling approach \cite{yannakakis2018ordinal}.

\section{Preference Learning for Modelling Motivation \label{sec:methods}}

Preference learning is a supervised machine learning technique \cite{furnkranz2011preference}, in which an algorithm learns to infer the preference relation between two variables. PL is a robust method which relies on relative associations instead of absolute values or class boundaries. PL applies a \emph{pairwise transformation} to the original dataset, yielding a representation of differences between feature vectors in the query \cite{furnkranz2003pairwise} which can be solved by a binary classifier. As an example, we observe the preference relation: $x_i \succ x_j \in X$ ($x_i$ is preferred over $x_j$) based on their associated output: $y_i > y_j$. Through the pairwise transformation two new features are created: $x'_1 = (x_i - x_j)$, associated with $y'_1=1$ and $x'_2 = (x_j - x_i)$, associated with $y'_2=-1$. This comparison between each pair of feature vectors provides $X' \subseteq X\cdot(X-1)$ new datapoints. $X'$ is a subset of all possible combinations because a clear preference relation can not always be inferred. 
{During the pairwise transformation, a hyperparameter is applied to control the preference threshold ($P_t$) during comparisons. For instance, with a $P_t=0.1$, only datapoints with a value difference above $10\%$ of the value of the second datapoint are considered (i.e. a clear preference). The purpose of this threshold is to counter the noise in the ground truth data which can skew modelling results. Although increasing $P_t$ eliminates insignificant differences and potential noise while increasing the accuracy of the model, sparsity of unique values in the ground truth can also lead to a rapid decrease in sample size, and thus hurt the robustness of the trained models.}

This study uses ranking \emph{Support Vector Machines} (SVM) \cite{joachims2002optimizing} as they are implemented in the \emph{Preference Learning Toolbox}\footnote{\url{http://plt.institutedigitalgames.com/}}\cite{farrugia2015preference} which is based on the LIBSVM library \cite{chang2011libsvm}. We choose SVMs in this initial study as they can yield robust models even with a limited amount of data and input features. SVMs were originally employed to solve classification tasks by maximizing the margins of a hyperplane separating the datapoints projected into a higher dimensional feature space \cite{vapnik1995statlearn} but were later adopted to solve PL tasks as well \cite{joachims2002optimizing}. We use both linear and non-linar SVMs with \emph{radial basis function} (RBF) kernels. Unlike linear SVMs, which aim for a linear separation between datapoints, RBF SVMs emphasize the local proximity of datapoints, fitting the maximum-margin hyperplane in a transformed feature space \cite{vapnik1995statlearn}. For tuning our algorithms, we rely on the $C$ regularization term which controls the trade-off between maximizing the margin and minimizing the classification error of the training set, and---in case of RBF kernels---the $\gamma$ hyperparameter, which controls how each comparison is weighted in the non-linear topology.

\section{The Game and the Data}

This section first presents the testbed game, and then discusses the collected data and steps taken to pre-process it.

\begin{figure}[!tb]
\centering
\includegraphics[width=1\columnwidth]{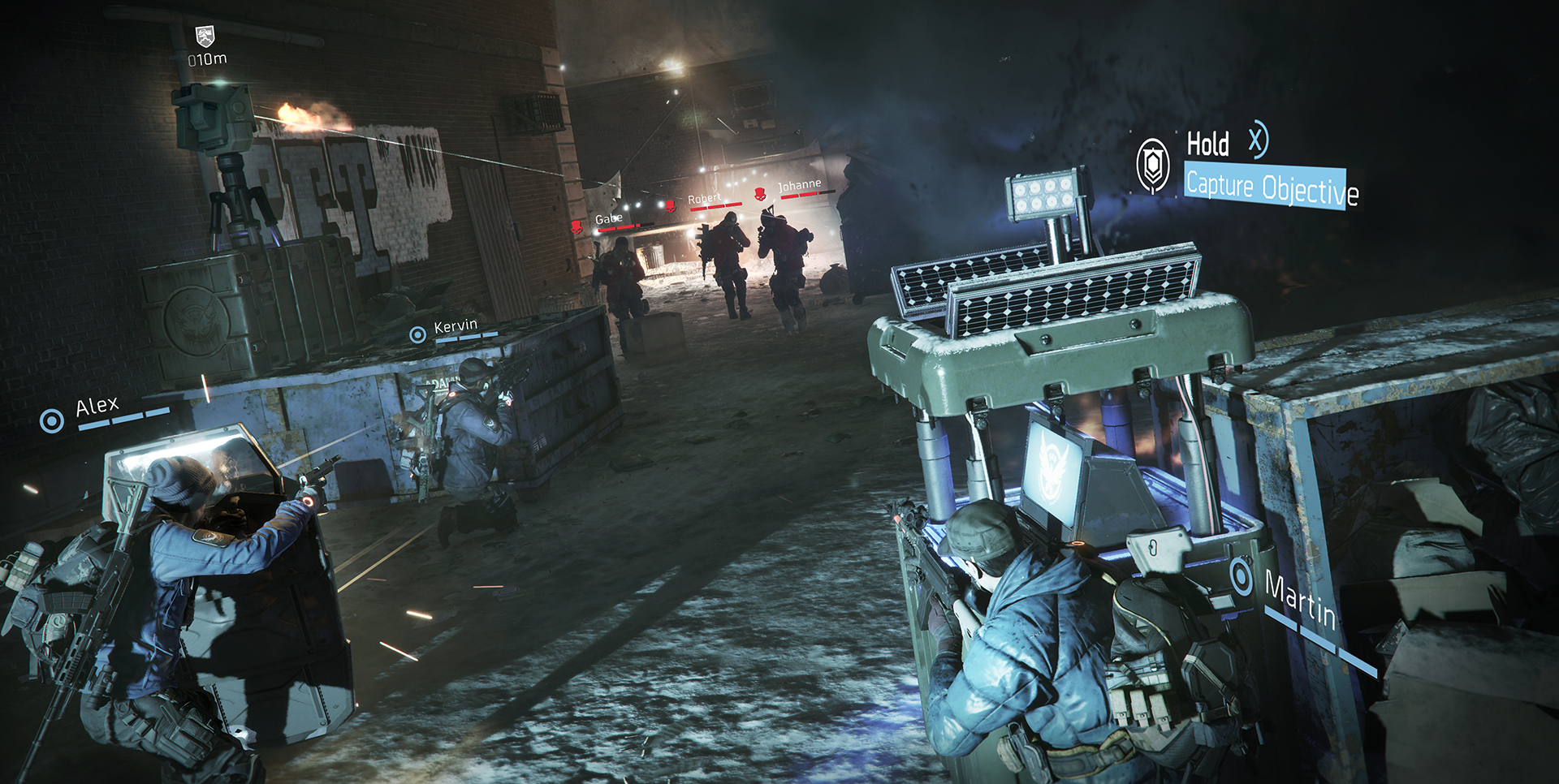}
\caption{An example of the gameplay of \emph{Tom Clancy's The Division} (Ubisoft, 2016). Image taken from: store.steampowered.com/app/365590. No copyright infringement intended.}
\label{fig:division}
\end{figure}

\subsection{Tom Clancy's The Division}

This study uses in-game behavioural data (player metrics) and survey questionnaire responses from players of \emph{Tom Clancy's The Division} (Ubisoft, 2016), hereafter The Division, collected between 2016 and 2018. The Division is an online multiplayer action-role playing game (Fig. \ref{fig:division}) set in a post-apocalyptic New York. Players, as government agents, have to work together (and against each other) to scavenge and investigate the fallen city. Gameplay relies on third-person, cover-based, tactical shooting combat mechanics.

The core of the game is a progression system, in which players gain new levels by participating in different in-game activities including story-focused and optional missions. Levels unlock new abilities and activities offer new equipment (e.g.~weapons, armour). The strength of a player can be measured by their level (up to 30) and by the quality of their equipment---expressed in \emph{Gear Score} points. In the player versus environment (PvE) sections of the game, players can group up and complete missions together. The game also features a competitive player versus player (PvP) area---called the \emph{Dark Zone}---which has its own progression system. In this special area players can still group up to complete missions for better equipment; however, they can also turn on each other and become \emph{Rogue} by killing other players and taking their rewards. At the maximum level (30), players can participate in \emph{Incursions}, which are particularly difficult missions for groups. Finally, downloadable content (DLC) adds new areas, equipment, and both PvE and PvP content to the game.

The game was well received (80/100 Metacritic score on consoles\footnote{\url{https://www.metacritic.com/game/xbox-one/tom-clancys-the-division}}) and was the best selling game of Ubisoft at the time of its release\footnote{\url{https://news.ubisoft.com/en-us/article/313497/the-division-sets-new-sales-records/}}. Due to a blend of PvP and PvE and support for different play styles and interaction modes, The Division is a rich and complex testbed for research on motivation. With the second instalment of The Division set to release on March 15, 2019\footnote{\url{https://news.ubisoft.com/en-us/article/342463/the-division-2-pc-features-specs-detailed/}}, a study on how the gameplay of The Division shapes the motivation of its players is both timely and relevant. 

\subsection{Dataset}

The collected data consists of aggregated information on the in-game activity of players over a long period of time and their corresponding UPEQ survey scores. These two types of data were collected independently, with the gameplay features recorded between 2016 and 2018 and the survey data collected through a web interface separately in 2018. As such, the survey data measures a general disposition of the players. 

The dataset consists of one datapoint per player. In  total, $443$ players participated in the above-mentioned data collection process. The dataset is cleaned of datapoints with missing values, corrupted entries, and outliers to prevent skewing any statistical analysis process. Extensive pruning was necessary due to outliers distorting the distribution of general game metrics (see Section \ref{sec:features}) and due to noise generated by the data logging service which inflated playtime. The clean dataset contains $298$ players. As part of a preliminary step in the PL task, the dataset is converted through a pairwise transformation, using the $P_t$ parameter (see Section \ref{sec:methods}). We report our findings with $P_t=0.1$, which preserves more than $60\%$ of the possible training and test samples. The transformation is applied within cross-validation folds in a 10-fold setup. After applying the appropriate $P_t$, on average we are left with $44\cdot10^3$ training and $536$ test samples for \emph{competence}; $39.5\cdot10^3$ training and $471$ test samples for \emph{autonomy}; $56.5\cdot10^3$ training and $681$ test samples for \emph{relatedness}; $52\cdot10^3$ training and $632$ test samples for \emph{presence}.

\subsection{Extracted Features \label{sec:features}}

Player behaviour is measured through $30$ high-level gameplay features. While most of these are simple aggregated \emph{game metrics} describing the time allocation and progression of the player, 4 of them are exclusive categories of distinct \emph{play styles} based on sequence-based profiling of the player's in-game activities \cite{makarovych2018like}. Additionally, the dataset contains $4$ Likert scores that represent the four \emph{motivation factors} of each player as measured by the UPEQ survey. The three types of data considered in this study as detailed as follows:

\subsubsection{Game Metrics} These features may relate to \emph{general playtime} (Days Played, Days in Groups, Days in the Dark Zone, 
Sessions, Playtime, Group Playtime, Dark Zone Playtime, Playtime as Rogue); \emph{completion} (Non-Daily Missions, Daily Missions, Side Missions, Days with Incursions, Incursions); \emph{progression} (Gear-Score, Dark Zone Rank, Level, Early Level 30\footnote{Early Level 30 shows whether the player reached the maximum level faster than the average.}, Reached Level 30); \emph{early gameplay} (Level Below 30, Early Playtime, Early Group Playtime, Early Dark Zone Playtime, Early Playtime as Rogue); and \emph{DLC gameplay} (Underground Playtime, Survival Playtime, Season-Pass).

\subsubsection{Player Types} The $4$ different player types are named \emph{Adventurer}, \emph{Elite}, \emph{PvE All-Rounder}, and \emph{Social Dark Zone Player}. These types have been derived through a traditional $k$-means clustering and qualitative interpretation of highly aggregated data. Adventurers are focused on solo main and side missions, Elites have the top gear score, PvE All-rounders engage in cooperative activities, while the Dark Zone Players prefer competitive activities. 

\subsubsection{Motivation Factors} UPEQ scores the four factors of motivation in the form of averaged Likert-scale values. While computing the mean of ordinal data can be problematic conceptually \cite{yannakakis2015ratings,yannakakis2018ordinal}, average survey scores are a wide-spread method of using Likert-like data as they can still show certain tendencies within the scores (e.g., a higher score is assumed to correspond to an overall more positive response). As mentioned in Section \ref{sec:ordinal}, we adopt a \emph{second-order} modelling approach \cite{yannakakis2018ordinal} and treat these scores as ordinal data through pairwise comparisons across all players.

\section{Models of Player Motivation\label{sec:results}}

This section presents the results of the machine learned models of player motivation based on different feature sets. As the extracted play styles are more complex descriptors of the player behaviour than the aggregated game metrics, we examine their capacity of predicting motivation independently (see Section \ref{sec:metrics} and \ref{sec:styles}) but also in fusion (Section \ref{sec:both}). 
{Results presented in this section measure model accuracies in terms of predicting the higher \emph{competence, autonomy, relatedness} and \emph{presence} in pairwise comparisons of players.} 
Before delving into the details of the obtained results, we outline the validation and parameter tuning process followed. 

\subsection{Validation and Parameter Tuning \label{sec:tuning}}
All PL models are validated with 10-fold cross validation. To prevent data leakage, the training and test folds are separated before the normalisation and pairwise transformation of the data. After the pairwise transformation of the dataset, the training and test inputs are $z$-normalised. To preserve the independence of the test set, we assume that it is drawn from the same distribution as the training set and apply z-normalisation to it based on the properties of the training set.

The best $C$ and $RBF\gamma$ hyperparameters are found in the $C \in \{1,2,3,4,5\}$ and $\gamma \in \{0.1, 0.5, 0.75, 1, 2\}$ parameter spaces. The best $\gamma$ is $0.5$ across all RBF models. For linear SVMs best performance is with $C=2$ for {competence}; $C=1$ for {autonomy}; $C=4$ {relatedness}; $C=3$ {presence}. For RBF SVMs best performance is with $C=3$ for {competence}; $C=4$ for {autonomy}; $C=4$ for {relatedness}; $C=2$ for {presence}.

\subsection{Models Based on Play Styles}
\label{sec:styles}

\begin{figure*}[!tb]
    \centering
    \subfloat[Play styles\label{fig:styles_results}]{
    \includegraphics[width=0.31\linewidth]{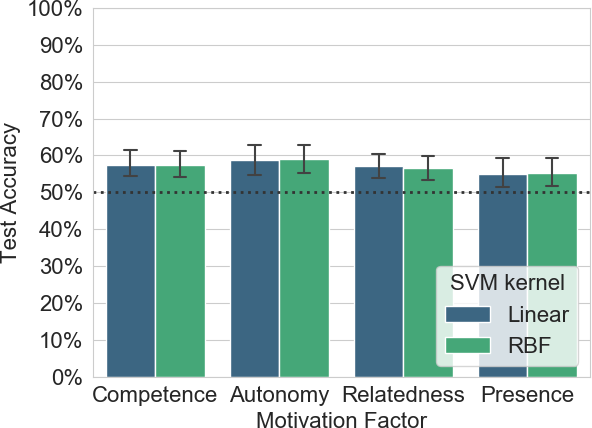}}
    \quad
    \subfloat[Game metrics\label{fig:metrics_results}]{
    \includegraphics[width=0.31\linewidth]{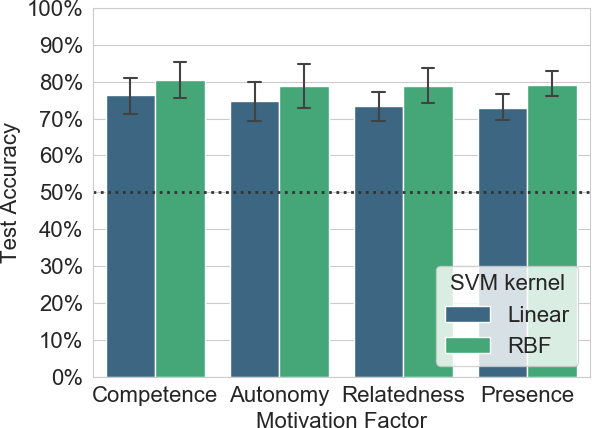}}
    \quad
    \subfloat[All features\label{fig:all_results}]{
    \includegraphics[width=0.31\linewidth]{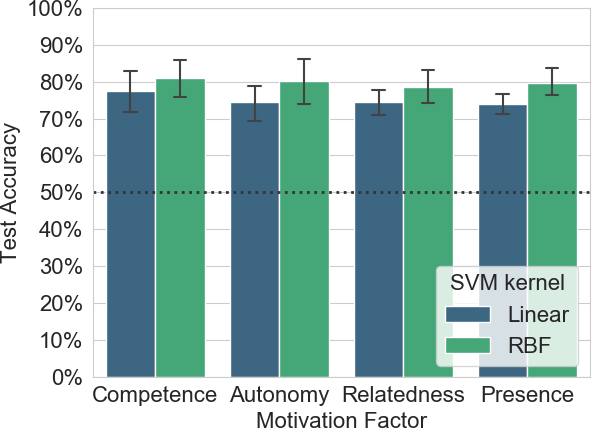}}
    
    \caption{Average accuracy of the best linear and RBF SVMs. The dotted line shows the baseline. The error bars represent $95\%$ confidence intervals.}
\end{figure*}

In the first round of experiments only the four play styles are used as input of the SVM model. Figure \ref{fig:styles_results} shows the final results. Despite the low dimensionality of the input feature set, both linear and non-linear models are able to surpass the $50\%$ baseline (with $57\%$ on average). Introducing an RBF kernel does not improve the average accuracy and results are comparable to linear models. Despite the low average performance, the best models reaching as high as $73.2\%$ on average 
(\emph{competence}: $73.2\%$; \emph{autonomy}: $70.68\%$; \emph{relatedness}: $65\%$; \emph{presence}: $71\%$)
Nevertheless, the low average score allows us to conclude that the four play styles on their own are not sufficient for building accurate models of motivation.

\subsection{Models Based on Game Metrics}
\label{sec:metrics}

In the second round of PL experiments models are trained only considering the $26$ game metrics as model inputs---as discussed in Section \ref{sec:features}. As evidenced by the bar plots of Figure \ref{fig:metrics_results}, game metrics are quite successful in predicting the reported motivation factors. In particular, linear SVM models are successful with an average accuracy of $74.31\%$ across all models while the best models for individual factors are performing at well above $80\%$ accuracy on certain folds (\emph{competence:} $89\%$; \emph{autonomy:} $85.25\%$; \emph{relatedness:} $83\%$; \emph{presence:} $86.5\%$)
Non-linear kernels further improve the model's performance to a $79.3\%$ accuracy on average across all models. The best individual models outperform the corresponding linear models reaching past $90\%$ accuracy (\emph{competence}: $94.3\%$; \emph{autonomy}: $93.55\%$; \emph{relatedness}: $92.6\%$; and \emph{presence}: $91.4\%$). Compared to linear models, RBF SVMs appear to be more robust across all motivation factors (approx. $5\%$ increase in accuracy). Unlike the poor performances of PL models based solely on play styles, models based on game metrics are very accurate and robust across all four factors. 

\subsection{Models Based on All Features}
\label{sec:both}

Even if under-performing on their own, play style information can enhance other PL models by adding domain-specific information. To test this assumption we pool play style profiles and game metrics together and train SVM models based on the combined input. Based on Fig.~\ref{fig:all_results}, including play styles has only marginal improvement, if any.
Linear Models perform $75\%$ on average across all tests with comparable peek performance to the models based on game metrics with an average best accuracy over $85\%$ (\emph{competence:} $89.4\%$, \emph{autonomy:} $84.33\%$, \emph{relatedness:} $83.3\%$, and \emph{presence:} $84.36\%$) for individual models.
RBF SVMs show a marginal improvement over those which are based on game metrics alone with $79.88\%$ average accuracy. The best models perform above $94\%$ on average (\emph{competence:} $94.3\%$; \emph{autonomy:} $94.5\%$; \emph{relatedness:} $92.3\%$; \emph{presence:} $92.33\%$). Similar to models based on game metrics, models based on all features show very similar performance across all four factors.

\subsection{Visualising the Motivation Models}

\begin{figure}[!tb]
\centering
\includegraphics[width=1\columnwidth]{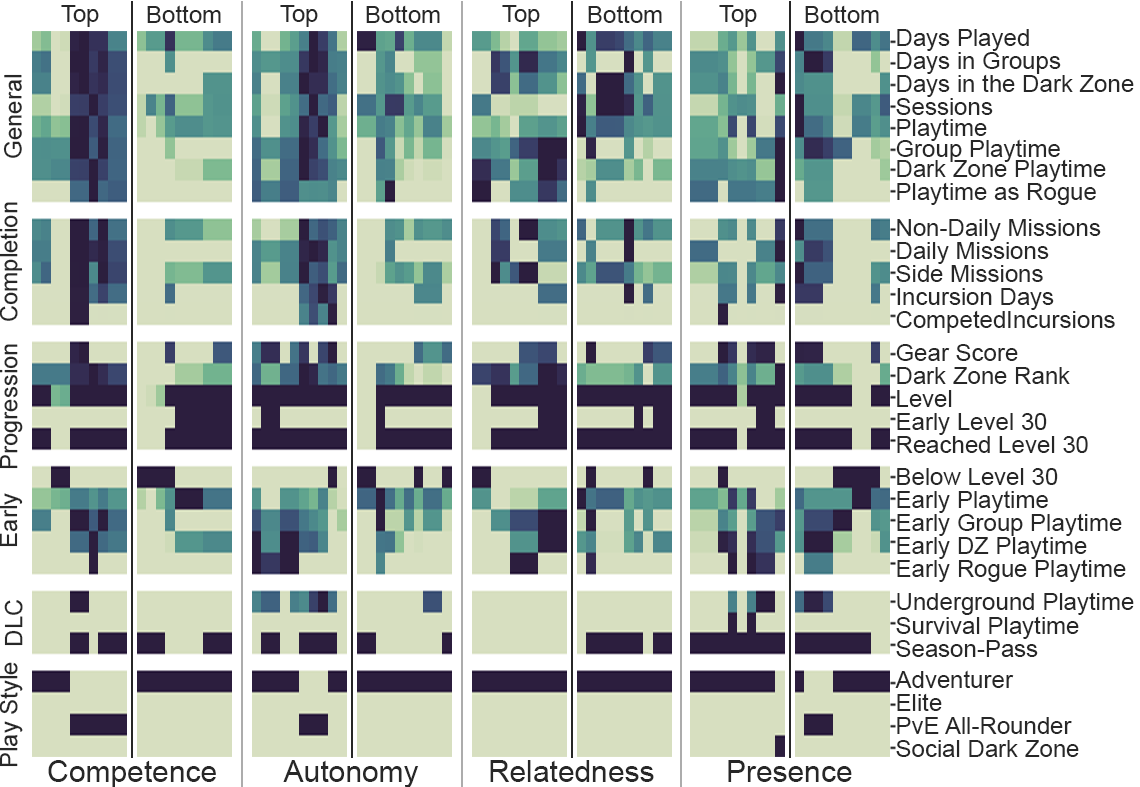}
\caption{Feature sets of the top and bottom 10 players ordered based on predicted motivation factors. Each column represents a player and each row is a different input feature. Darker cells show high values (normalised to each value range).}
\label{fig:motivation_factors}
\end{figure}

To demonstrate the applicability of the models for game design, we re-examine the players' features ordered by the predictions of each model. Using the predictions of the best SVM models, we create a global ordering of players on each of the four motivation factors and visualise the feature sets for the top 10 and the bottom 10 players (Fig. \ref{fig:motivation_factors}).

As evidenced by Figure \ref{fig:motivation_factors}, players with predicted high competence have more experience with the game and have spent more time playing with various systems of the game in general. The high concentration of \emph{Early Level 30} in the low competence group suggests that these players are rushing through the content. 
While there is disparity in terms of general playtime in the high competence group, high completion rates and/or \emph{Dark Zone} and \emph{Group Playtime} (coupled with no \emph{Early Level 30}) indicate that \emph{competence} aligns with a steady progression through the game.

In predicted \emph{autonomy}, a similar effect can be observed in terms general playtime and completion. Additionally to high completion and progression rates, PvE and PvP success from an early level (Completed Incursions, and Dark Zone and Rogue Playtime) as well as DLC playtime also seem to be a deciding factor between high and low perceived \emph{autonomy}.

In terms of \emph{relatedness} the picture is more fuzzy. While both the top and bottom players share many similar features, the most prominent pattern is the high level of social interaction within the top group either as pro-social (e.g.\emph{Group Playtime}) or antagonistic (e.g \emph{Playtime as Rogue}).

Finally, ordering by perceived \emph{presence} shows a seemingly uniform distribution across different features with no discernible patterns between top and bottom players. Even so, the best models could still predict the higher ranking players in pairwise comparisons with $92.33\%$ within given folds.

These observations largely match theories of SDT, further validating both the predictive models and the UPEQ questionnaire \cite{azadvar2018upeq}. Across all factors, the most divergent features between top and bottom ranked players are associated with general playtime with an emphasis on mission completion, PvP activity, and playing with others in a group. Even in cases where no obvious linear relationship exists between individual features and motivation factors, non-linear PL techniques prove to be efficient methods for predicting motivation and offer an insightful qualitative tool for game design.

\section{Discussion}

This study applied preference learning via SVMs to model psychological constructs of the self determination theory and demonstrated the models' supreme efficiency and robustness in predicting players' motivation factors based solely on their in-game data. The SVM models can be applicable directly in a games user research context to predict factors of players' motivation based on aggregated data from multiple play sessions, which can be the subject of further qualitative analysis. In this study, we collected and processed the data of $298$ players of \emph{Tom Clancy's The Division} and defined motivation quantitatively using the UPEQ questionnaire. We converted the Likert scores to ordinal values using a \emph{second-order} data processing approach \cite{yannakakis2018ordinal} that yielded data corpora of over $4\cdot10^4$ 
samples, representing the differences between players' UPEQ Likert scores in pairs. The predicted pairwise relationships between players in terms of \emph{competence}, \emph{autonomy}, \emph{relatedness} and \emph{presence} with more than $80\%$ accuracy on average, while the best models reach accuracies of at least $92\%$.

The presented research offers insights for game industry professionals and stakeholders, who aim to leverage and enhance the positive psychological effects of gameplay but also for researchers in affective computing and games user research. The paper offers a novel approach that utilises a machine (preference) learning approach to model psychological constructs such as motivation by treating the subjectively defined notions as an ordinal phenomenon. To the best of our knowledge, this is the first study which attempted to quantitatively model and predict constructs of SDT based on player behaviour in a commercial-standard game and with that level of success.

A core limitation of this work is posed by the processing of the motivation factors (i.e., the ground truth). In particular, we opted for a simple experimental design and took the aggregated UPEQ scores at face value; aggregating Likert scale scores in this way, however, poses a number of methodological issues. The most important of these limitations is the processing of ordinal values as interval data \cite{yannakakis2015ratings}. As the distance between ordinal points is arbitrary, their mean value is not necessarily meaningful and introduces subjective reporting bias to the measurements \cite{yannakakis2017ordinal}. To address this issue, future research should focus on new ways of transforming Likert-like data to preference relations. One possible approach would be to observe how an individual player's baseline shifts through the survey, scoring the four observed factors depending on significant changes in the responses. 

Another limitation is posed by the nature of the dataset. As the dataset contains aggregated gameplay data, we cannot observe clearly how each player's motivation changes over time and thus the relationship between the player's ``lifetime'' and a one-time post-experience survey is rather broad. Although this type of data is easier to handle in a qualitative analysis, machine learning models are able to handle larger datasets with a higher dimensional feature space. Future research could focus on collecting a new dataset, recording multiple sessions and questionnaires per participant, and creating models which can predict temporal changes in the player's motivation.

It is important to note that the ordinal data processing method we propose in this paper yields large datasets which are generated through the pairwise comparison of all players. As such, the method is viable in current game development settings since it only requires a small sample of player experience annotations---such as those usually available in quality assurance departments of game studios. Although SVMs already showcased high levels of efficiency in predicting motivation on sets of several thousand datapoints, it can be argued that they might not be as robust when faced with much larger datasets of that type. In such instances alternative methods derived from \emph{deep preference learning} \cite{martinez2014deep,martinez2013learning} are directly applicable to the modelling task.

\section{Conclusion}

This paper attempted to find a computational mapping between gameplay data and aspects of player motivation, measured through the \emph{Ubisoft Perceived Experience Questionnaire}. In this way, the \emph{competence}, \emph{autonomy}, \emph{relatedness} and \emph{presence} of almost 300 players of \emph{Tom Clancy's The Division} was collected, as well as these players' in-game data. Experiments in this paper explored the degree to which such data can be a powerful predictor of players' survey responses. For that purpose, survey scores were converted to ordinal values and preference learning was applied to create efficient and robust support vector machine models of the four factors of motivation. The ordinal machine learning approach proved extremely successful in predicting such complex psychological constructs by eliminating the reporting bias of the questionnaires. Core findings in this paper suggest that not only is it possible to infer the mapping between high-level gameplay metrics and survey-based annotations of complex emotional and cognitive states, but the inferred models have predictive capacities that reach certainty levels.

\bibliography{bibliography.bib}
\bibliographystyle{IEEEtran}
\end{document}